\documentclass{amsart}
\usepackage{amsmath}
\usepackage{graphicx}
\usepackage{amsfonts}
\usepackage{amssymb}
\usepackage{subfigure}

\theoremstyle{plain}

\numberwithin{equation}{section}

\catcode`@=11
\def\plainvspace@{\def\vspace##1{\noalign{\vskip##1}}}
\def\plainLet@{\relax\iffalse{\fi\let\\=\cr\iffalse}\fi}
\def\aligned{\,\vcenter\bgroup\plainvspace@\plainLet@\openup\jot\m@th\ialign
  \bgroup \strut\hfil$\displaystyle{##}$&$\displaystyle{{}##}$\hfil\crcr}
\def\endaligned{\crcr\egroup\egroup}
\def\matrix{\,\vcenter\bgroup\plainLet@\plainvspace@
    \normalbaselines
  \m@th\ialign\bgroup\hfil$##$\hfil&&\quad\hfil$##$\hfil\crcr
    \mathstrut\crcr\noalign{\kern-\baselineskip}}
\def\endmatrix{\crcr\mathstrut\crcr\noalign{\kern-\baselineskip}\egroup
                \egroup\,}

\def\cases{\left\{\,\vcenter\bgroup\plainvspace@
     \normalbaselines\openup\jot\m@th
      \plainLet@\ialign\bgroup$\displaystyle{##}$\hfil&\quad$\displaystyle{{}##}$\hfil\crcr
      \mathstrut\crcr\noalign{\kern-\baselineskip}}
\def\endcases{\endmatrix\right.}
\begin{document}
\title{Segmentation for radar images based on active contour }
\author   {
Meijun Zhu \& Pengfei Zhang }

\begin{abstract}
We exam various geometric active contour methods for radar image
segmentation. Due to special properties of radar images, we
propose our new  model based on modified Chan-Vese functional. Our
method is  efficient in separating non-meteorological noises from
meteorological images.

\end{abstract}

 \maketitle

\section{Radar image processing}

Weather radar data quality control is extremely important for
meteorological and hydrological applications.  For weather radars,
scatterers in the atmosphere are not only meteorological particles
like cloud, rain drops, snowflakes, and hails, but also
non-meteorological particles such as chaff, insects, and birds.
For radar meteorologists it is a major issue and challenge to
design numerical scheme which can  automatically and subjectively
distinguish meteorological echoes from non-meteorological echoes.
Non-meteorological echoes can contaminate radar reflectivity and
Doppler velocity measurements, and subsequently  cause errors and
uncertainty on radar data applications in  quantitative
precipitation estimation,  as well as in  assimilation in
numerical model for weather prediction. Automatic detection of
tornado or mesocyclone vortex among meteorological echoes is
certainly  another big challenge and has great potential in
improving severe weather forecast and saving human life.

\begin{figure}[bh]
\label{fig0} \centering \subfigure[\it The famous May 03, 1999
Tornado (Moore, Oklahoma) radar image]{\includegraphics[width=160
pt, height=140 pt]{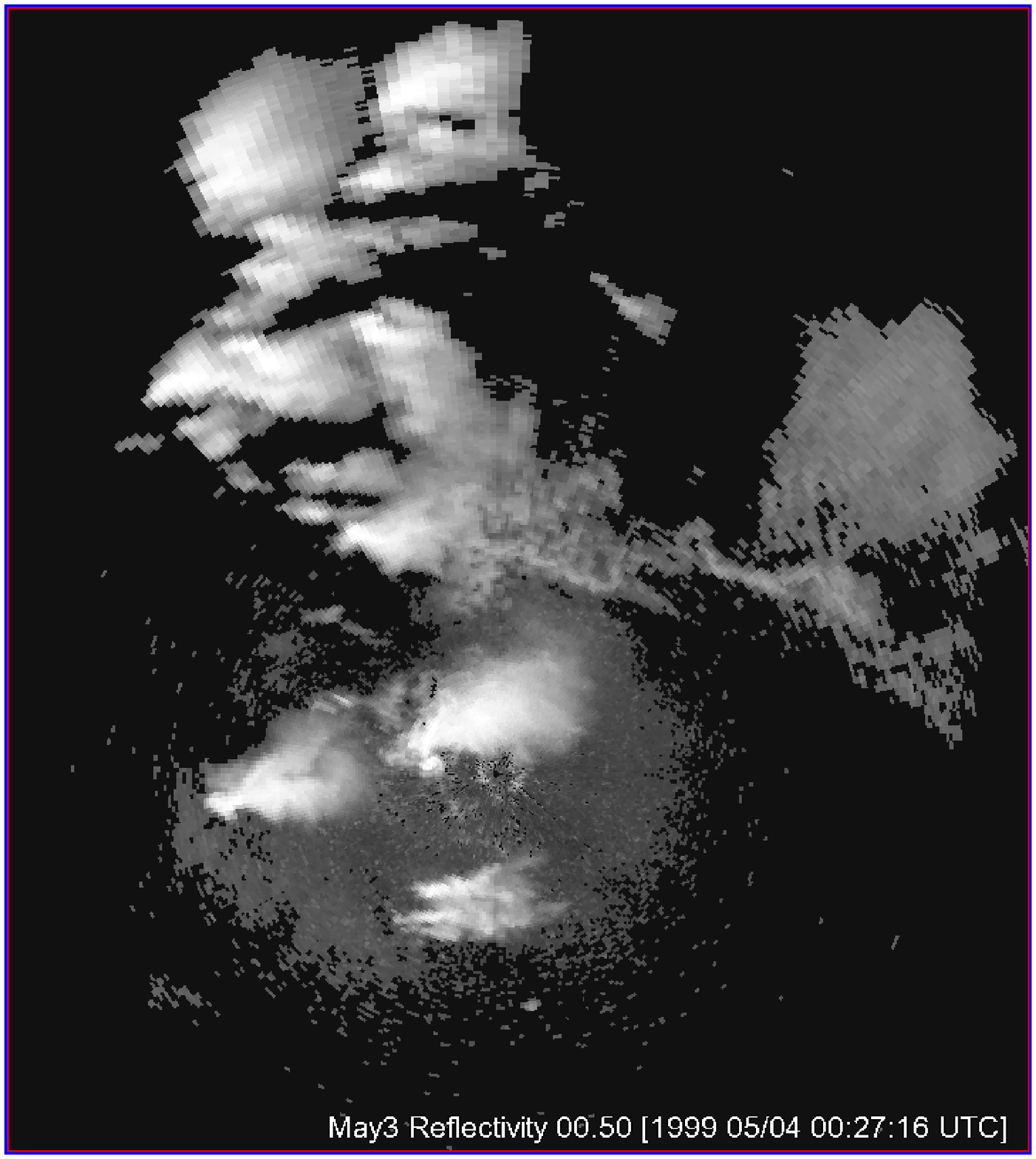}}
 \subfigure[Noised storm image: radar noise is embedded in storm image]{\includegraphics[width=160 pt, height=140
pt]{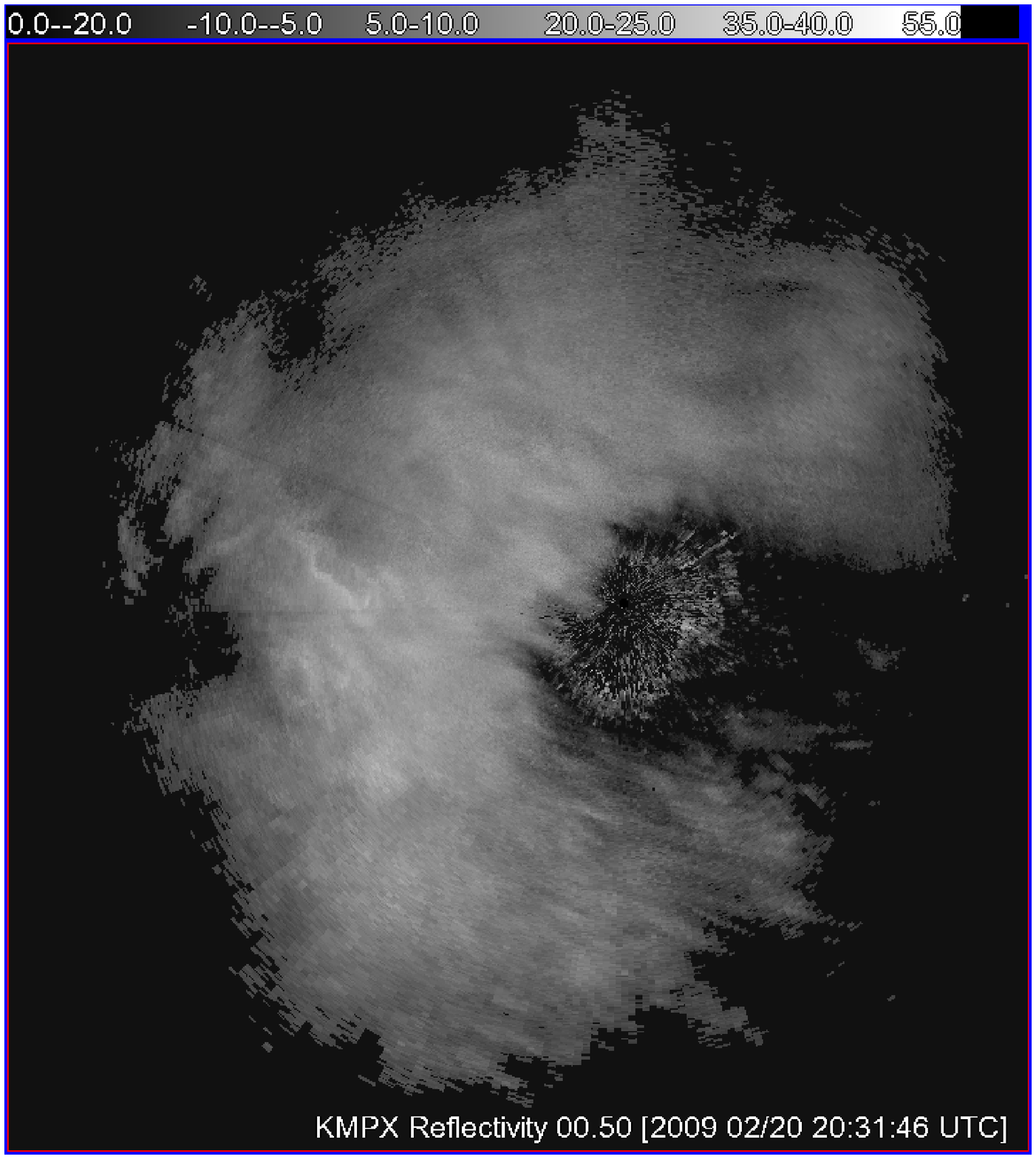}} \caption{ \bf Challenges in radar images}
\end{figure}

For different echoes, their graphic properties such as pattern,
intensity and texture, are different. See, for example, fig.
\ref{fig0}. Such differences enable us to design new  active
contour model  to automatically extract
 the most distinguishing   graphic properties from echoes,
and  to  segment them easily from other noises. Our methods shall
also be very useful in automatic detection of tornadic supercell,
as well as in  storm classifications and tracking in the future
study.

%
%

\section{Active contour}
Active contour is the procedure that we deform a given curve  so
that a given  functional of the curve will achieve its local
minimal value. This method is wildly used recently in computer
vision in seeking the edges or contours of given images. See, for
example, Mumford and Shah \cite{MS}, Kass, Witkin and Terzopoulos
\cite{KWT}, Caselles, Kimmel and Sapiro \cite{CKS}, and Chan and
Vese \cite{CV}.

Let $u_0(x, y): \Omega \to R$ be the  gray level function of a
given image. If $u_0$ is smooth,  then the edge of the image are
those points $(x, y)$ where   $|\nabla u_0|$ is relatively large.

The geometric contour and snake models aim to detect edge
automatically based on the size of $|\nabla u_0|$.  Let $C(s): [0,
L] \to \mathbb{R}^2$ be a closed curve, where $s$ is its arc
length parameter. One can introduce an edge-detector function $g:
\Omega \to R_+$ so that $g(z) \to 0$ as $z\to \infty$. A typical
example of such function is given by
$$
g(z)=\frac 1{1+z^2}.$$  We define the  energy functional of $C$ by
\begin{equation}
I_1(C):=\int_0^L g(\nabla u_0(C(s))) ds, \label{m-1}
\end{equation}
then to find the edge of  image  $u_0$ can be reduced to seek the
local minimal for $I_1$ (the geometric active contour model
\cite{CKS}):
\begin{equation}
I_1(edge)=inf_{C} I_1(C). \label{mm-1}
\end{equation}

The snake model \cite{ KWT}  is to introduce,  for a parameterized
curve $C(p): [0, 1] \to \mathbb{R}^2$, the following energy
functional :
\begin{equation}I_2(C)=\alpha\int_0^1|C'(p)|^2 dp+\beta
\int_0^1|C''(p)| dp-\lambda \int_0^1|\nabla u_0(C(p))|^2 dp,
\label{m-2}
\end{equation}
where $\alpha, \beta, \lambda$ are all positive parameters. The
first two terms represent the internal energy of the image, which
usually are used to smooth the curve; The third term represents
the external energy, serving as the indicator for edge. The edge
of the image then can be found by minimizing $I_2$:
\begin{equation}
I_2(edge)=\inf_{C} I_2(C). \label{mm-2}
\end{equation}

%
%
%
%
%
%
%
%

\bigskip

To automatically detect the edge via an iteration scheme, one
introduces a family of curves $C(p, t): [0, 1] \times [0, \infty)
\to  \mathbb{R}^2$ and the deformation path.
For example, for active contour model (\ref{mm-1})  the curve
evolution (gradient flow equation) is given by
\begin{equation}
C_t=(kg-\nabla g\cdot \mathbf{N})\mathbf{N}, \label{1-1}
\end{equation}
 where $k$ is the curvature function and $\mathbf{N}$ is
the inner unit normal vector of curve $C(p, t).$

Numerically, such iteration can be realized via the powerful level
set method of Osher and Sethian \cite{OS}.  Embed $C(p, t)$ as a
nodal line of a smooth function $\Phi(x, y, t)$: $C=\{(x, y, t) \
: \ \Phi(x, y, t)=0\}$. From $\partial_t \Phi(C, t)=0$, we are led
to evolve $\Phi$ by
\begin{equation}
\cases & \frac{\partial \Phi}{\partial t}=g(|\nabla u_0|)
div (\frac{\nabla \Phi}{|\nabla \Phi|}) |\nabla \Phi|+<\nabla g, \nabla \Phi>\\
&\Phi(x, y, 0)= \Phi_0(x, y),
\endcases
 \label{1-2}
\end{equation}
where $\Phi_0(x, y)$ is the initial level set function. In
practice, one can choose $\Phi_0(x, y)$ to be a signed distance
function to a given initial curve $C(p, 0)$. Fig. 2 (b), fig. 3
(b), and fig. 4 (b) show the result for image processing based on
such scheme.

\medskip

If the given image $u_0(x, y)$ is not smooth, the edge of the
image is not well defined based on the  derivative of the gray
level function. The human being's perspective for the edge of a
non smooth image basically is to identify the boundary of
different groups. To identify such boundary, one can use
Chan-Vese energy \cite {CV}:
\begin{align}
I_3(C, c_1, c_2):&=\int_{inside (C)}|u_0-c_1|^2dx dy+
\int_{outside
(C)}|u_0-c_2|^2dx dy \nonumber\\
&+ \mu\cdot \mbox(length (C))+\nu\cdot
\mbox(Area(inside(C))),\label{mm-3}
\end{align}
where $c_1$, $c_2$ are constants to be adjusted in iteration,
$\mu$ and $\nu$ are fixed parameters. The last two terms are
smoothing terms. The edge is again sought by minimizing $ I_3(C,
c_1, c_2)$: \begin{equation} I_3(edge, c_{1, *}, c_{2,
*})=\inf_{C, c_1, c_2} I_3(C, c_1, c_2).  \label{mm-3-1}
\end{equation}
Again, numerically level set method  can be used  for such
deformation. Introduce  the Heaviside function and its derivative
$$
H(z)=\cases &1, \ \ \ \mbox{if} \ \ z\ge 0\\
&0, \ \ \ \mbox{if} \ \ z\ge 0,
\endcases
\ \ \ \ \ \ \delta(z)=\frac{d}{d z} H(z).
$$
Embedding  $C(p, t)$ as a nodal line of a smooth function $\Phi(x,
y, t)$: $C=\{(x, y, t) \ : \ \Phi(x, y, t)=0\}$, we can re-write
the energy functional $I_3(C, c_1, c_2)$ as
\begin{align}
J_3(\Phi, c_1, c_2):&=\int_{\Omega}|u_0-c_1|^2 H(\Phi(x, y))dx dy+
\int_{\Omega}|u_0-c_2|^2(1-H(\Phi(x, y)))dx dy \nonumber\\
&+ \mu  \int_{\Omega}\delta(\Phi(x, y))|\nabla \Phi(x, y)|dx
dy+\nu \int_{\Omega} H(\Phi(x, y))dx dy.\label{mm-3-3}
\end{align}
For fixed $\Phi$, minimizing $J_3(\Phi, c_1, c_2)$ with respect to
$c_i$ yields
$$
\cases &c_1(\Phi)= \mbox{average} (u_0) \ \mbox {in} \ \{\Phi<
0\}\\
&c_2(\Phi)= \mbox{average} (u_0) \ \mbox {in} \ \{\Phi\ge 0\}.
\endcases
$$
Once  $c_1$ and $c_2$ are fixed, we minimize $J_3$ via  deforming
$\Phi$ along the gradient  direction:
\begin{equation}
\cases & \frac{\partial \Phi}{\partial t}= \delta(\Phi) \big \{\mu
\cdot div(\frac{\nabla \Phi}{|\nabla\Phi|}) -\nu
-(u_0-c_1(\Phi))^2+(u_0-c_2(\Phi))^2 \big\} \ \ \  \mbox{in}  \ \ \ (0, \infty)\times \Omega,\\
& \Phi(x, y, 0)  =\Phi_0(x, y) \ \ \ \ \ \mbox{in} \ \ \ \Omega,\\
& \frac {\delta (\Phi)}{|\nabla \Phi|} \frac {\partial
\Phi}{\partial n}=0 \ \ \ \ \ \ \ \mbox{on }\ \ \ \partial \Omega
\endcases\label{1-5}
\end{equation}
where $\Phi_0(x, y)$ is  a signed distance function to a given
initial curve $C(p, 0)$. Fig. 2 (c), fig. 3 (c), and fig.  4 (c)
show the result for image processing based on such scheme.

\medskip

\section{Modified model}
As being pointed out in \cite{CV}, The Chan-Vese model is
originated in the Mumform-Shah model \cite{MS}:
\begin{align*}
F^{MS}(u, C)=\mu \cdot length(C)&+ \lambda \int_\Omega |u_0(x, y)-
u(x, y)|^2 dx dy\\
&+\int_{\Omega \setminus C} |\nabla u(x, y)|^2 dx dy,
\end{align*}
where $u_0: \overline \Omega \to \mathbb{R}$ is the  given image,
$K\subset \Omega$ is the curve that we will deform. The sharp
boundary of $u_0$, where $|\nabla u_0|$ is large or discontinuous
can be detected via minimizing $F^{MS}(u, C)$. Roughly speaking,
if we replace $u$ by a constant function in the above, we can see
the prototype model similar to that of  Chan-Vese.  These models
are all viewed as minimal partition problems.

We observe that radar noises usually have relatively low intensity
to severe storms, and  radar signals for storm are usually uniform
in certain region. To segment  the severe storm from usual radar
noises, we introduce the following modified Chan-Vese functional:
\begin{align}
I_4(C, c):&=\int_{inside (C)}|u_0-\alpha \cdot M|^2dx dy+
\int_{outside
(C)}|u_0-c|^2dx dy \nonumber\\
&+ \mu\cdot \mbox(length (C))+\nu\cdot
\mbox(Area(inside(C))),\label{mm-4}
\end{align}
where $M=\max_{x \in \Omega} u_0(x)$, $c$ is a constant to be
adjusted in iteration, $\alpha$, $\mu$ and $\nu$ are fixed
parameters. In practice, parameter $\alpha$ can be determined by
comparing the maximal intensity near radar and the maximal
intensity of storm.


Embedding  $C(p, t)$ as a nodal line of a smooth function $\Phi(x,
y, t)$: $C=\{(x, y, t) \ : \ \Phi(x, y, t)=0\}$, we can re-write
the energy functional $I_4(C, c)$ as
\begin{align}
J_4(\Phi, c):&=\int_{\Omega}|u_0-\alpha \cdot M|^2 H(\Phi(x, y))dx
dy+
\int_{\Omega}|u_0-c|^2(1-H(\Phi(x, y)))dx dy \nonumber\\
&+ \mu  \int_{\Omega}\delta(\Phi(x, y))|\nabla \Phi(x, y)|dx
dy+\nu \int_{\Omega} H(\Phi(x, y))dx dy.\label{mm-4-4}
\end{align}
For fixed $\Phi$, minimizing $J_4(\Phi, c)$ with respect to $c$
yields
$$
c(\Phi)= \mbox{average} (u_0) \ \mbox {in} \ \{\Phi\ge 0\}.
$$

To derive the first variation of the functional, we consider
slightly regularized version of functions $H_\epsilon$ and
$H'_{\epsilon}=\delta_{\epsilon}$ such that $H_\epsilon \in
C^(\overline{\Omega})$, $H_\epsilon \to H$ and $\delta_\epsilon
\to \delta$, and the modified functional:
\begin{align}
J_{4,\epsilon}(\Phi, c):&=\int_{\Omega}|u_0-\alpha \cdot M|^2
H_\epsilon(\Phi(x, y))dx dy+
\int_{\Omega}|u_0-c(\Phi)|^2(1-H_\epsilon(\Phi(x, y)))dx dy \nonumber\\
&+ \mu  \int_{\Omega}\delta_\epsilon(\Phi(x, y))|\nabla \Phi(x,
y)|dx dy+\nu \int_{\Omega} H_\epsilon(\Phi(x, y))dx
dy.\label{mm-4-4-1}
\end{align}
Its first variation is
\begin{align*}
<\delta J_{4,\epsilon}, \psi> & =\int_{\Omega}(u_0-\alpha \cdot
M)^2 \delta_\epsilon(\Phi(x, y)) \psi dx dy-
\int_{\Omega}(u_0-c(\Phi))^2 \delta_\epsilon(\Phi(x, y)) \psi dx dy \nonumber\\
&+ \mu  \int_{\Omega}\delta_\epsilon(\Phi(x, y))\frac {\nabla
\Phi}{|\nabla \Phi(x, y)|} \nabla \psi dx dy+\nu \int_{\Omega}
\delta_\epsilon(\Phi(x, y)) \psi dx dy.
\end{align*}
Thus we can minimize $J_{4, \epsilon}$ via deforming $\Phi$ along
its gradient direction:
\begin{equation}
\cases &\frac{\partial \Phi}{\partial t}= \delta_\epsilon(\Phi)
\big \{\mu \cdot div(\frac{\nabla \Phi}{|\nabla\Phi|}) -\nu -
(u_0-\alpha
\cdot M)^2+(u_0-c(\Phi))^2 \big\} \ \ \  \ \ \mbox{in}  \ \ (0, \infty)\times\Omega,\\
 & \Phi(x, y, 0)  =\Phi_0(x, y) \ \ \ \ \ \ \ \mbox{in} \ \ \ \Omega,\\
 & \frac {\delta_\epsilon(\Phi)}{|\nabla \Phi|} \frac {\partial \Phi}{\partial n}=0 \ \ \ \ \ \ \ \mbox{on} \ \ \ \partial \Omega.
\endcases \label{2-1}
\end{equation}


\section{Experimental Results and Discussion}

An image of reflectivity factor field of tornadic supercells,
observed by a S-band weather radar near Oklahoma City, Oklahoma,
is presented in Fig.1(a). The violent tornadoes generated from the
weather system ripped through Oklahoma and Kansas and killed 48
people while demolishing houses and business. It had caused at
least $\$500$ M in property damage
(http://www.nssl.noaa.gov/headlines/outbreak.shtml).

  We first use geodesic active contour model
(\ref{1-2}) for the image. It results in catching all boundaries,
including radar noises (fig. 2 (b)). We then apply standard
Chan-Vese model (\ref{1-5}) with $\mu=5$, $\nu=0$ (fig. 2(c)).
Still it keeps almost all boundaries from radar noises. We finally
apply our model (\ref{2-1})  with $\alpha=0.7$ in fig. 2 (d). The
result shows that almost all radar noises are successfully
skipped.

Fig. 3 (a) usually is a challenge radar image for processing. The
radar noises are embedded in storm image (in fact, radar is
underneath the cloud). Geodesic active contour is very sensitive
to the initial curve. It usually contracts curve. Fig. 3 (b) is a
failure via  geodesic active contour. Fig. 3 (c) is the result
using Chan-Vese model (choose $\mu=5$, $\nu=0$); Fig. 3 (d) is
based on our model (\ref{2-1}) (with $\alpha=0.4$). There is no
big difference between fig. 3 (c) and fig. 3 (d).

Fig. 4 (a) is another radar image with radar noise separated from
storm.  Chan-Vese can not remove radar noise completely. Our model
with suitable parameter ( $\alpha=.6$) works fine.

\medskip

Finally, we compare the results using
 Chan-Vese model and our model with different parameters.

 First we consider  Chan-Vese model  with different parameters:
\begin{equation}
\cases & \frac{\partial \Phi}{\partial t}= \delta(\Phi) \big \{\mu
\cdot div(\frac{\nabla \Phi}{|\nabla\Phi|}) -\nu
-\lambda (u_0-c_1(\Phi))^2+(u_0-c_2(\Phi))^2 \big\}\\
& \Phi(x, y, 0)  =\phi_0(x, y), \endcases\label{5-5}
\end{equation}
where $\lambda$ is a positive parameter.  The results using
Chan-Vese model with different $\lambda$ are presented in fig. 5.

Next we compare the results using our model with different
$\alpha$ in fig. 6. It can be seen that for $\alpha$ in certain
range, our results are relatively stable. Therefore, for different
$\alpha$, if we let
$$
c_\alpha^{out}(\Phi)=\mbox{average} (u_0) \ \mbox{in} \
\{\Phi<0\},$$ we can develop a program which can automatically
determine which $\alpha$ we shall choose  based on the changing of
$c_\alpha^{out}(\Phi)$ as $\alpha$ changes.

\section{Conclusion}
We compare various models and their applications to the
segmentation of radar images. We propose our new model. Our method
is more efficient in outlining more severe storm images, and
skipping the usual radar noises.

\medskip
\noindent{\bf ACKNOWLEDGMENT. } M. Zhu  is partially supported by
the NSF grant DMS-0604169.


\begin{figure}[bh]
\label{fig1} \centering \subfigure[\it Original Tornado
image]{\includegraphics[width=160 pt, height=100
pt]{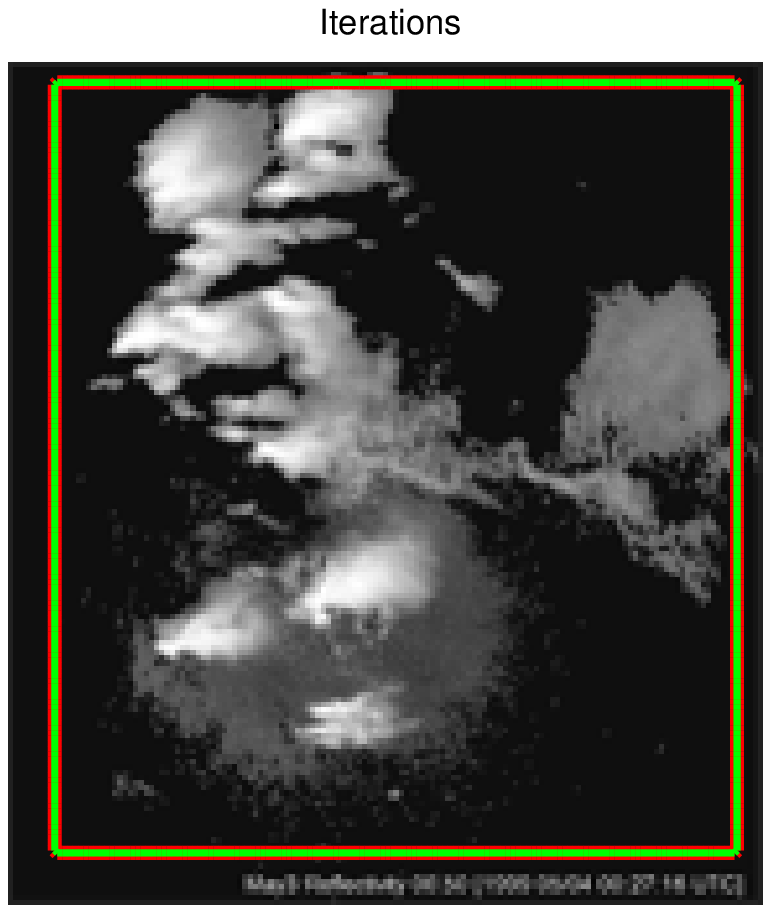}}
 \subfigure[Geodesic active contour: failed]{\includegraphics[width=160 pt, height=100
pt]{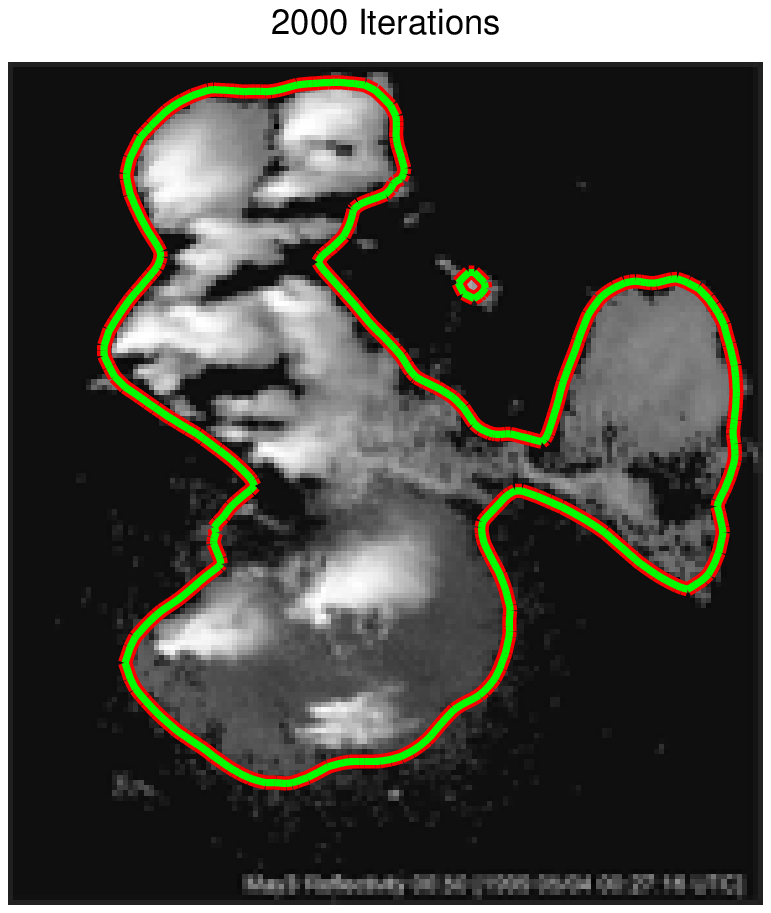}}

\centering \subfigure[ Chan-Vese model: Catch almost all
boundaries, majority of radar noise is
kept]{\includegraphics[width=160 pt, height=100 pt]{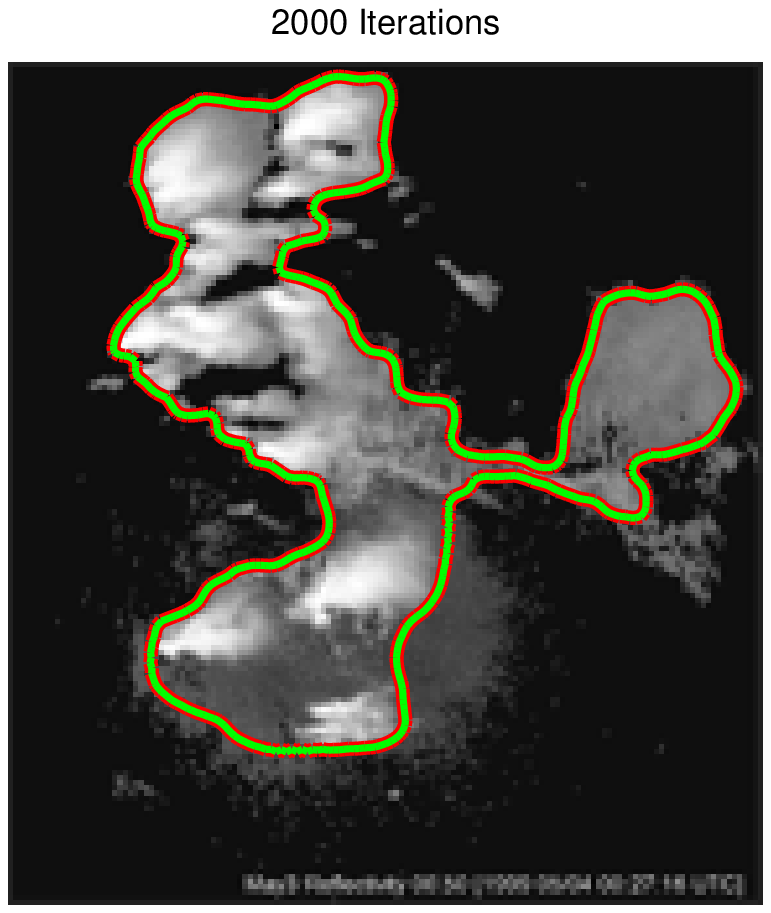}}
\subfigure[\it Our model: Catch serious storm, skip radar
noises]{\includegraphics[width=160 pt, height=100 pt]{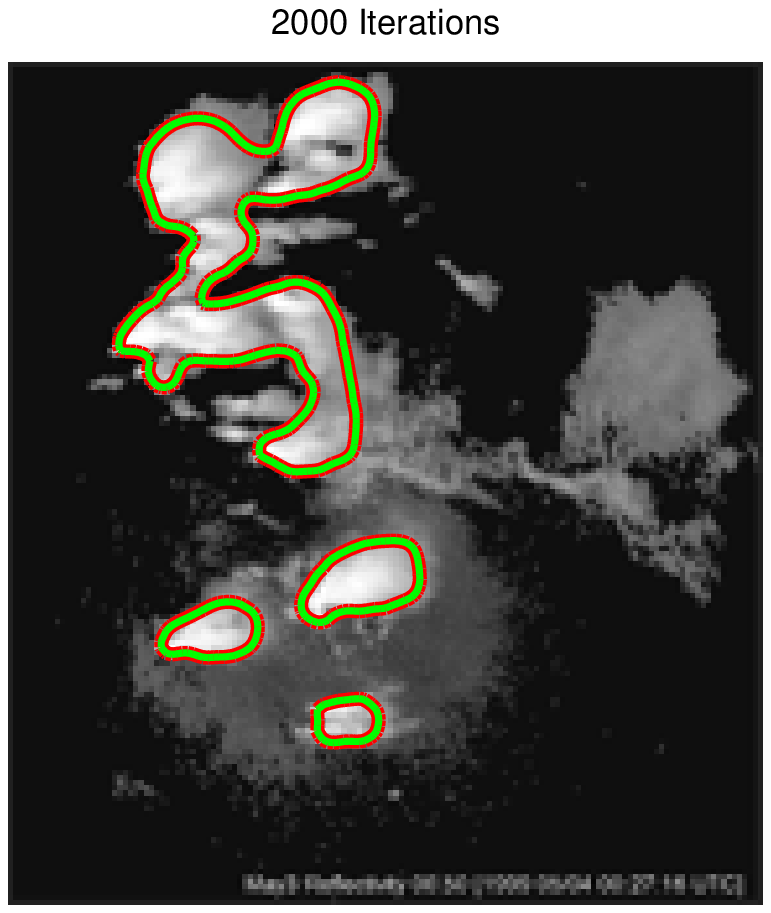}}
\caption{ \bf Comparison for different method}
\end{figure}

\begin{figure}[bh]
\label{fig2} \centering \subfigure[\it Radar noises are embedded
in storm image]{\includegraphics[width=160 pt, height=100
pt]{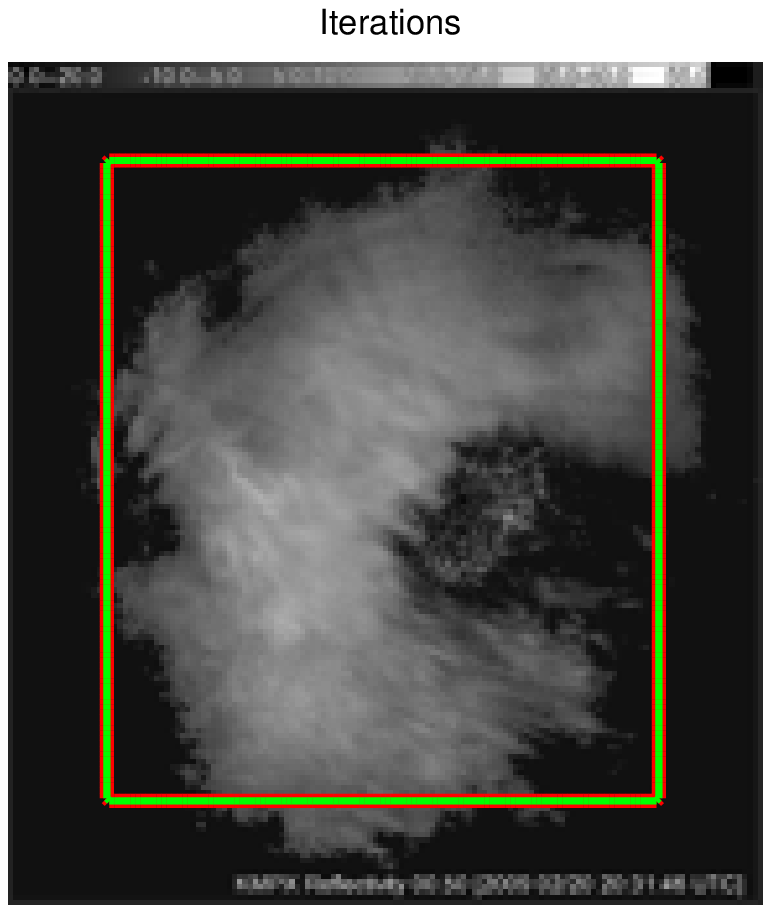}}
 \subfigure[Geodesic active contour: Can not expand]{\includegraphics[width=160 pt, height=100
pt]{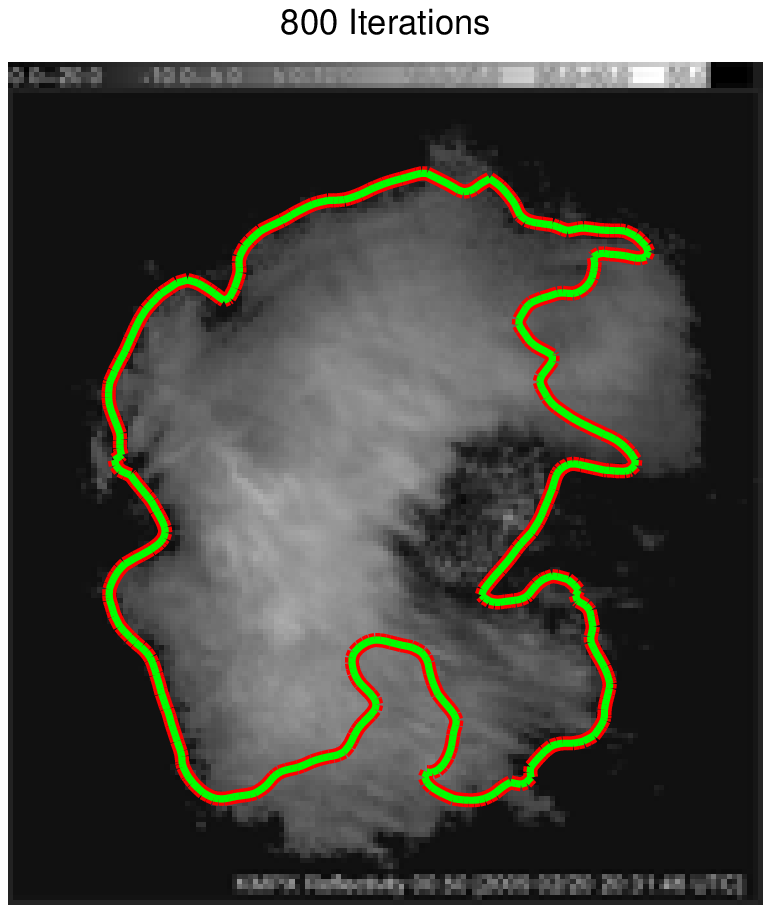}}

\centering \subfigure[ Chan-Vese model: skip radar noises
]{\includegraphics[width=160 pt, height=100 pt]{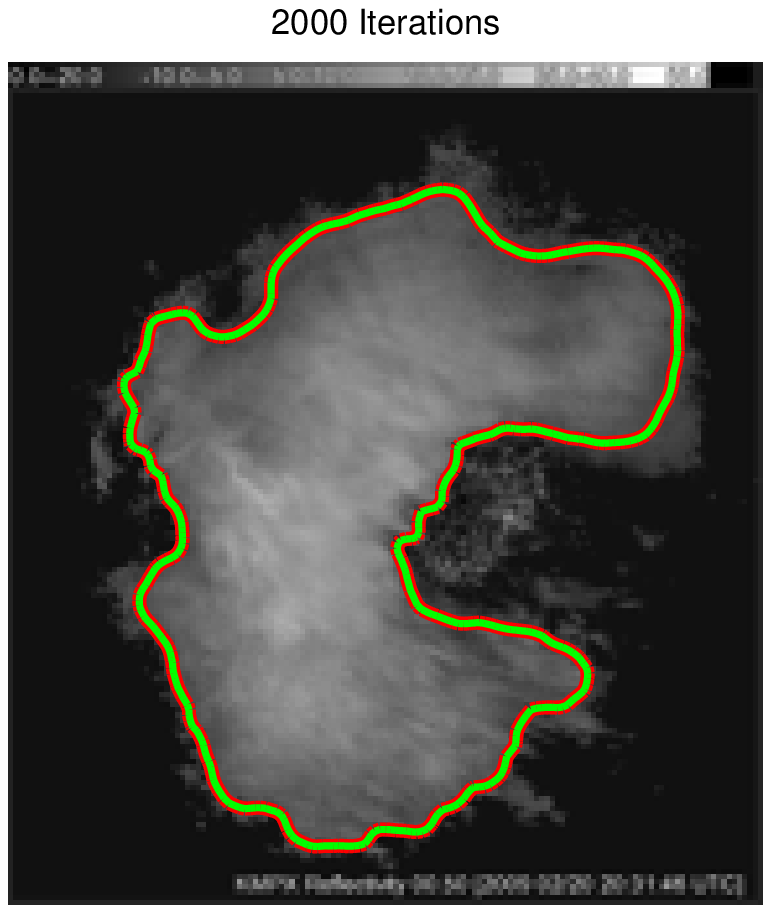}}
\subfigure[\it Our model: Catch serious storm, skip radar
noise]{\includegraphics[width=160 pt, height=100
pt]{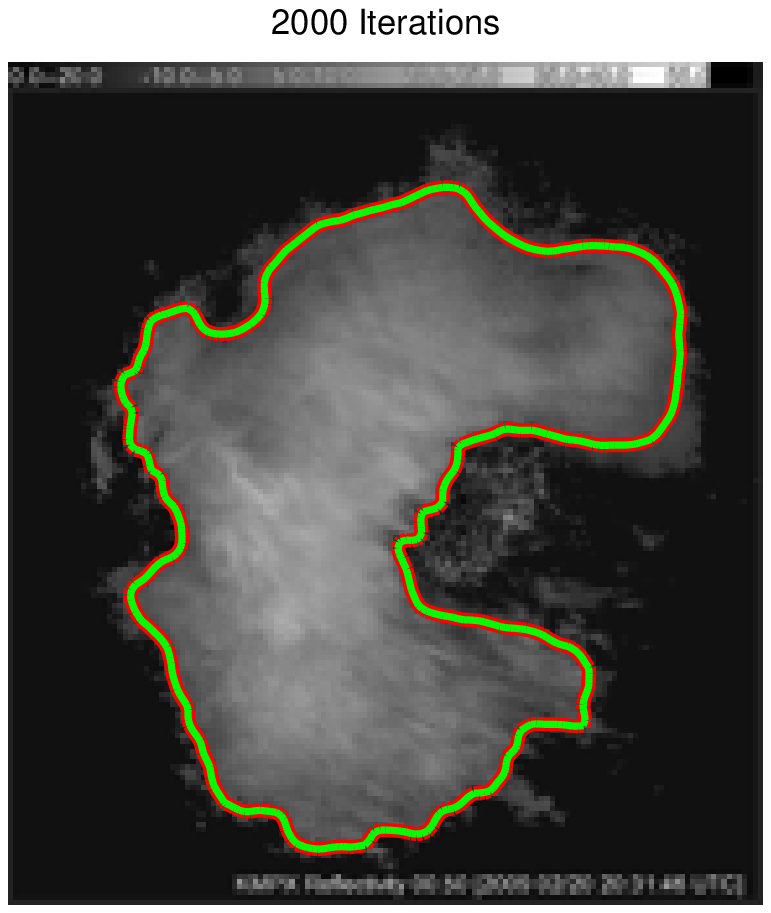}} \caption{ \bf Comparison for different
method}
\end{figure}

\begin{figure}[bh]
\label{fig3} \centering \subfigure[\it Radar noises are away from
the storm]{\includegraphics[width=160 pt, height=100
pt]{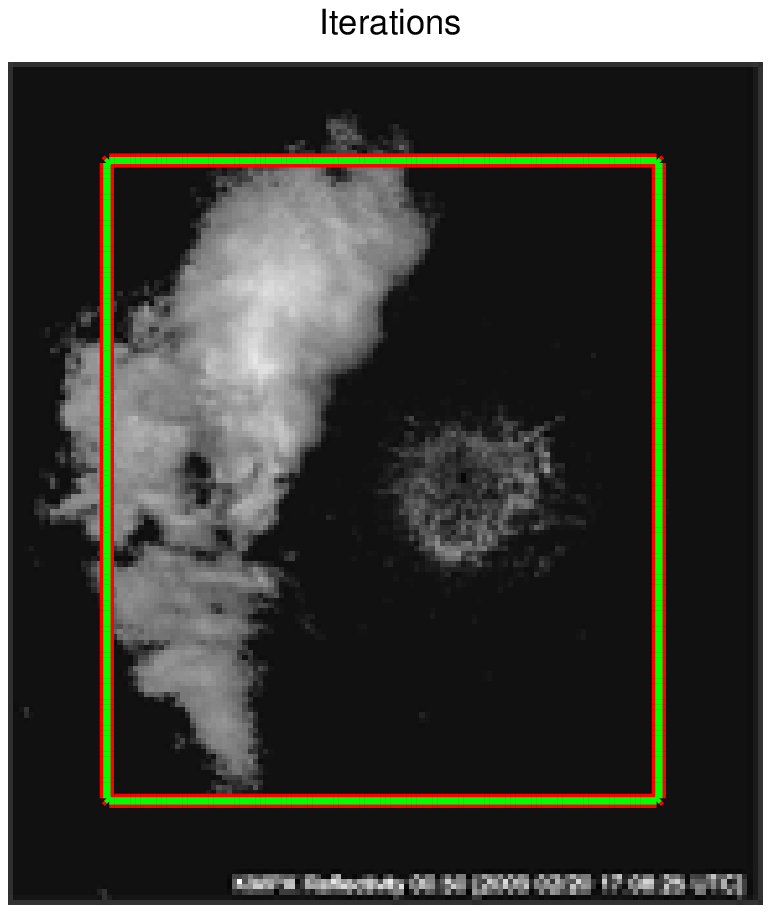}}
 \subfigure[Geodesic active contour: Catch all boundary]{\includegraphics[width=160 pt, height=100
pt]{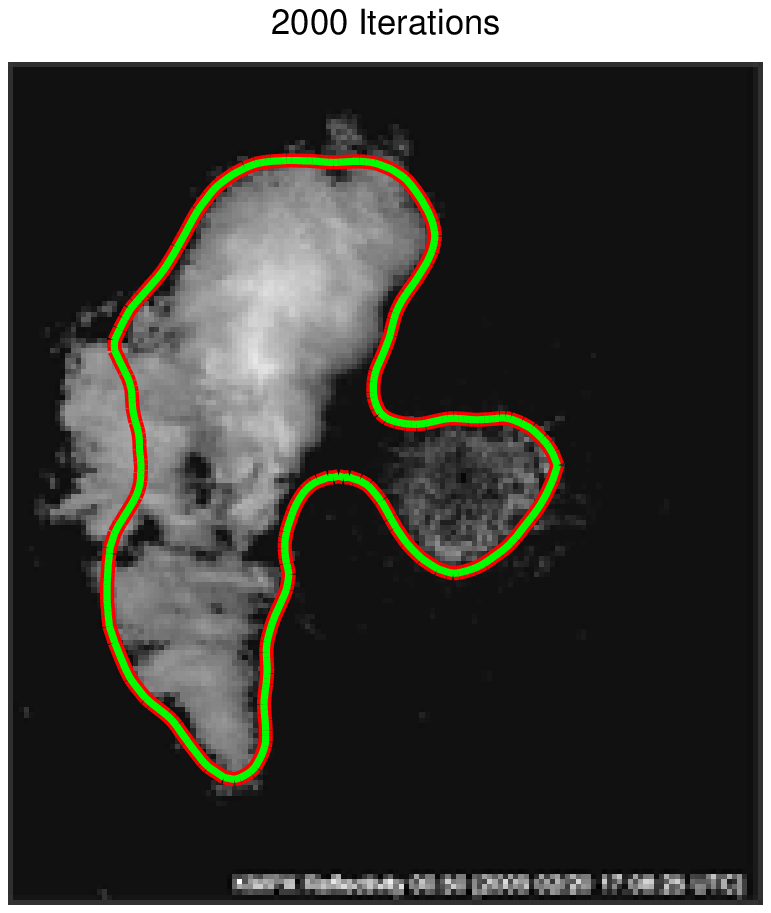}}

\centering \subfigure[ Chan-Vese model: Catch almost boundary,
small part of radar noise is kept ]{\includegraphics[width=160 pt,
height=100 pt]{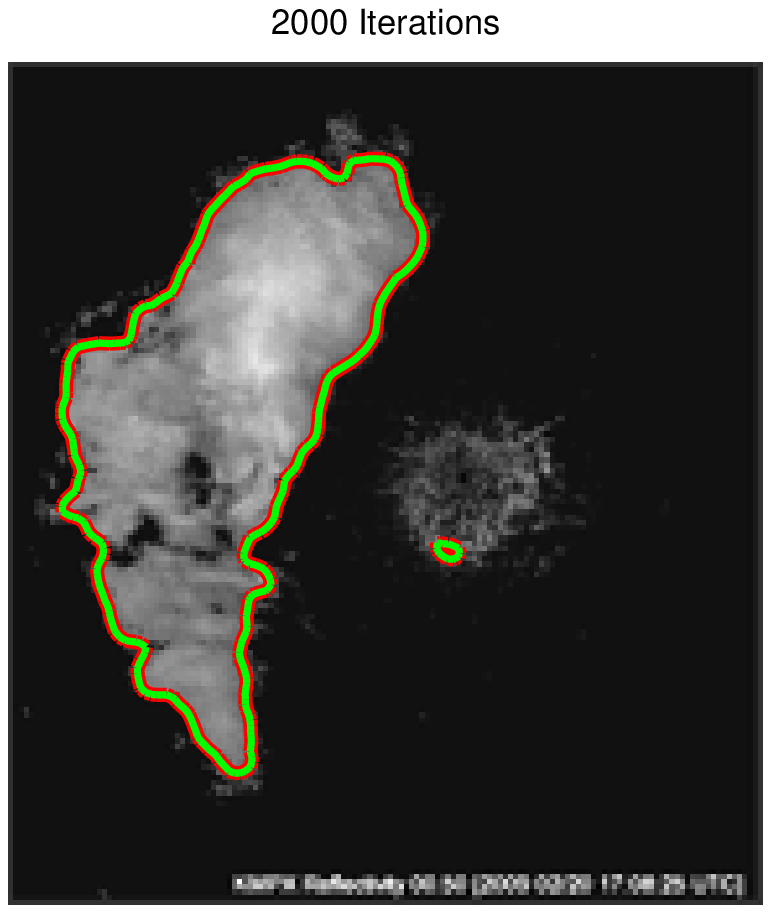}} \subfigure[\it Our model: Catch serious
storm, skip radar noise]{\includegraphics[width=160 pt, height=100
pt]{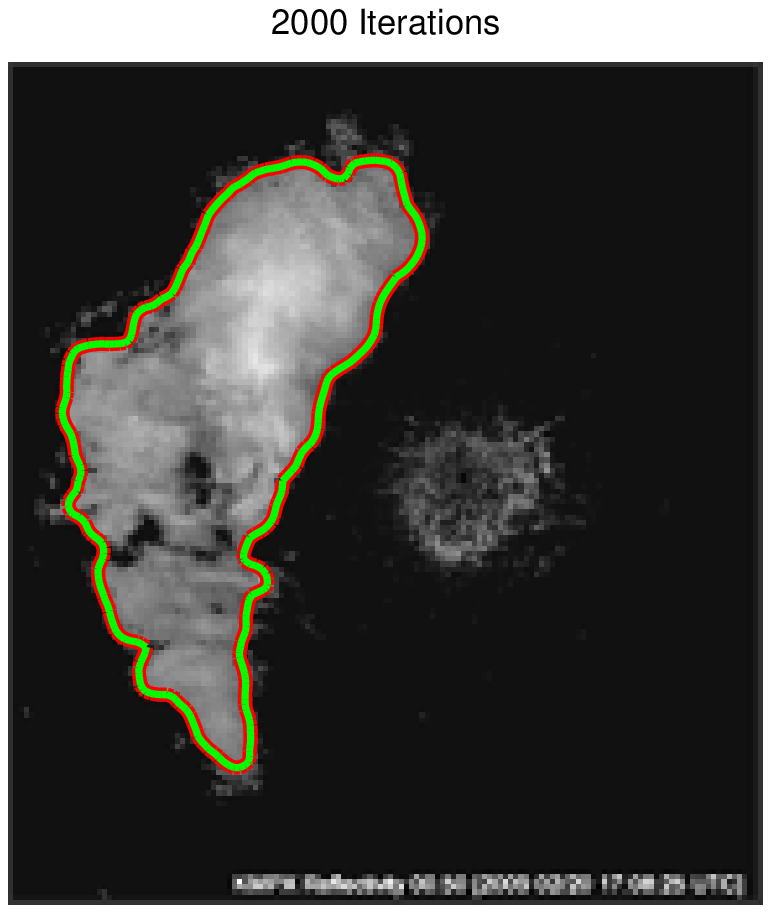}} \caption{ \bf Comparison for different method}
\end{figure}

\begin{figure}[bh]
\label{fig4} \centering \subfigure[\it Original Tornado
image]{\includegraphics[width=160 pt, height=100
pt]{Zgray3_1.eps}}
 \subfigure[$\lambda=2$]{\includegraphics[width=160 pt, height=100
pt]{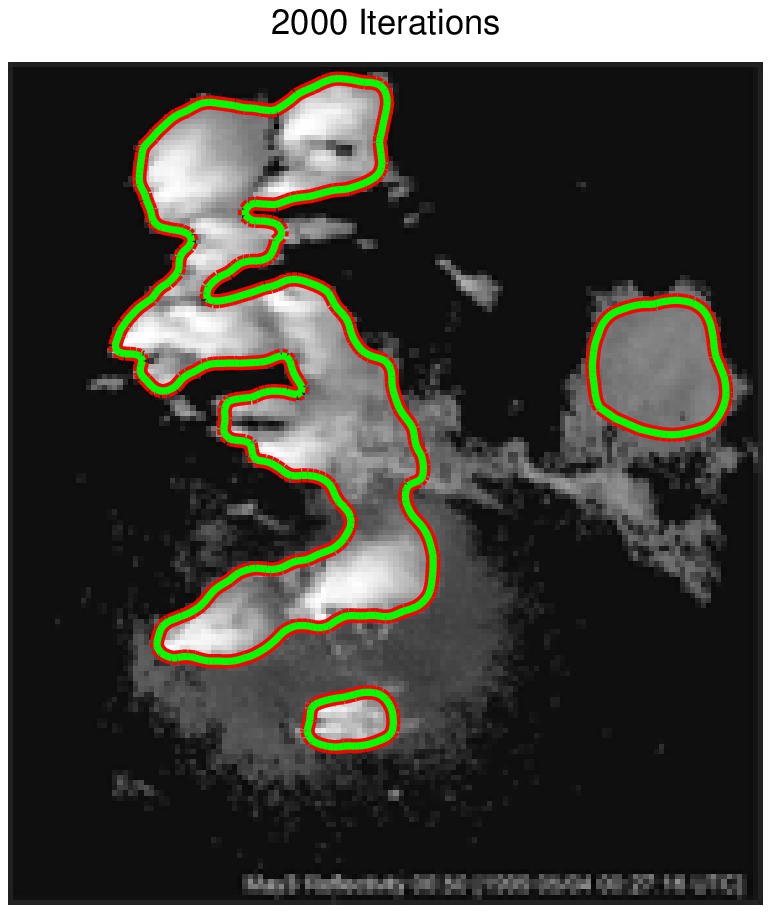}}

\centering \subfigure[ $\lambda=3$]{\includegraphics[width=160 pt,
height=100 pt]{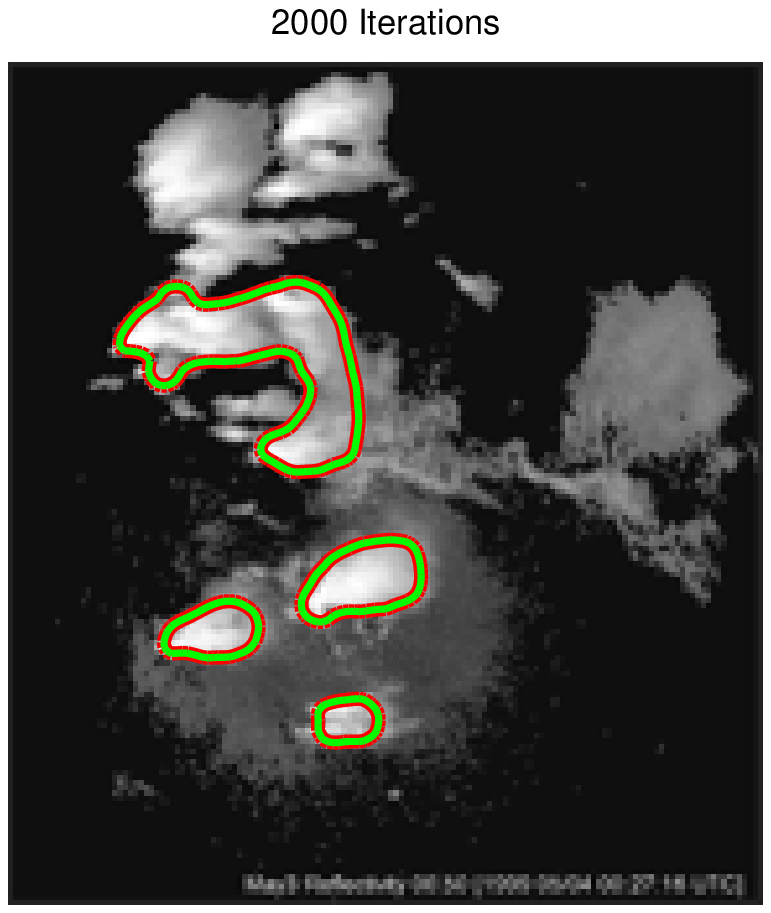}}
\subfigure[$\lambda=4$]{\includegraphics[width=160 pt, height=100
pt]{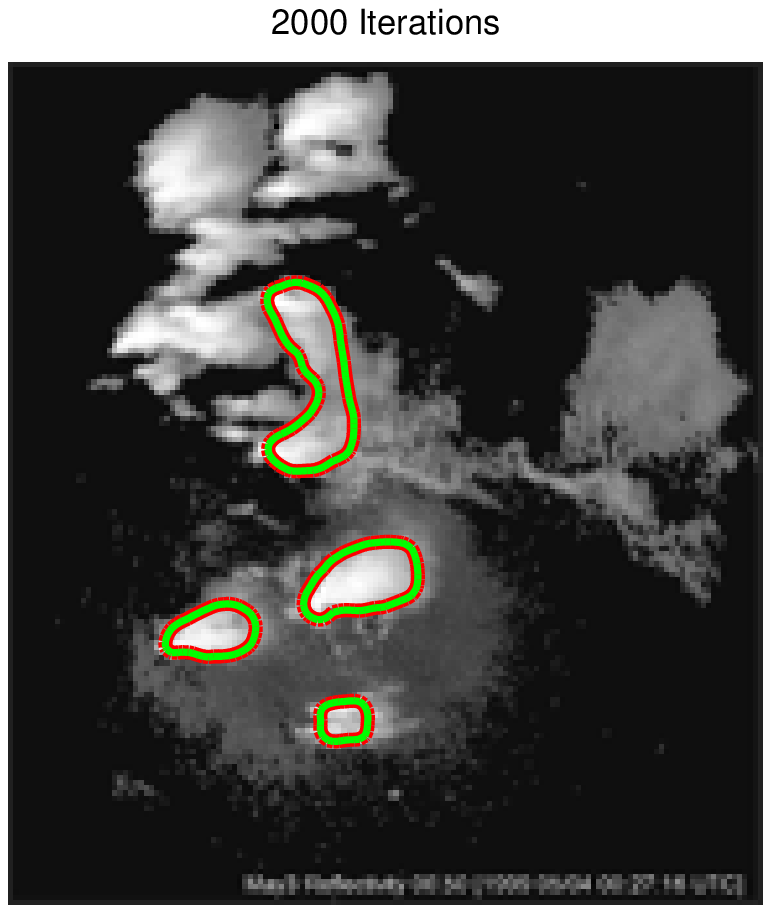}} \caption{ \bf Chan-Vese model with different
$\lambda$}
\end{figure}

\begin{figure}[bh]
\label{fig5} \centering \subfigure[\it Original Tornado
image]{\includegraphics[width=160 pt, height=100
pt]{Zgray3_1.eps}}
 \subfigure[$\alpha=0.3$]{\includegraphics[width=160 pt, height=100
pt]{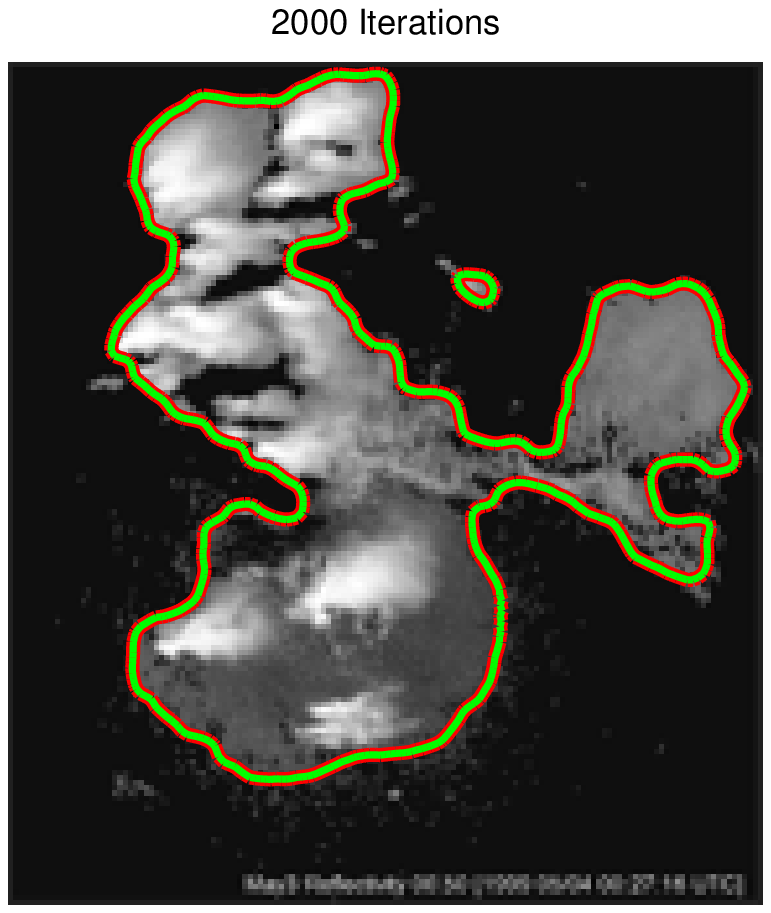}}

\centering \subfigure[ $\alpha=0.7$]{\includegraphics[width=160
pt, height=100 pt]{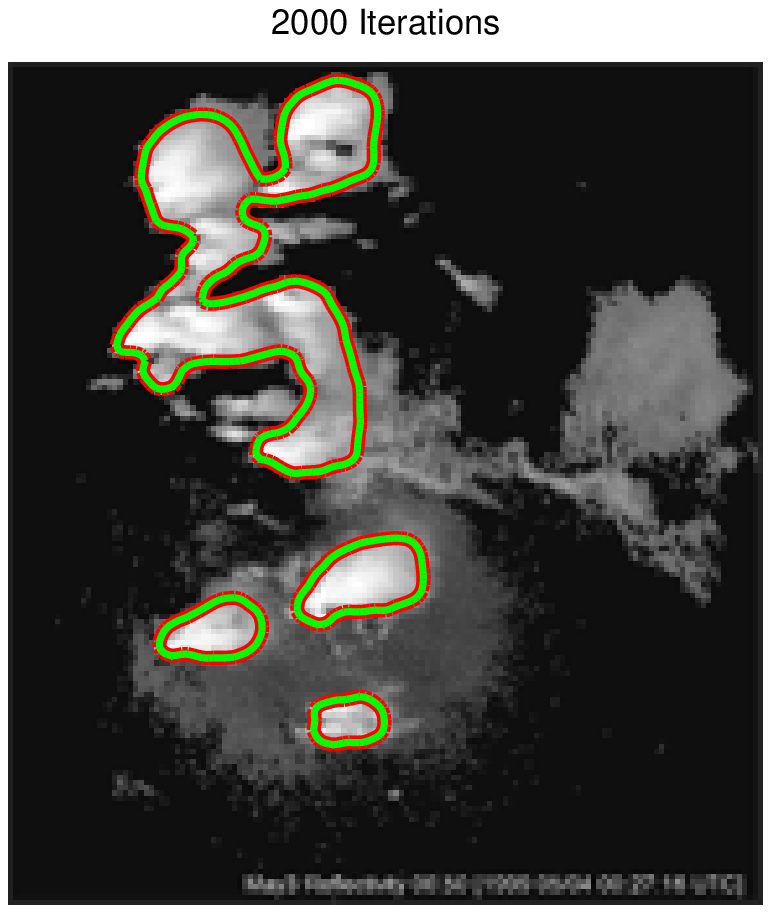}}
\subfigure[$\alpha=0.9$]{\includegraphics[width=160 pt, height=100
pt]{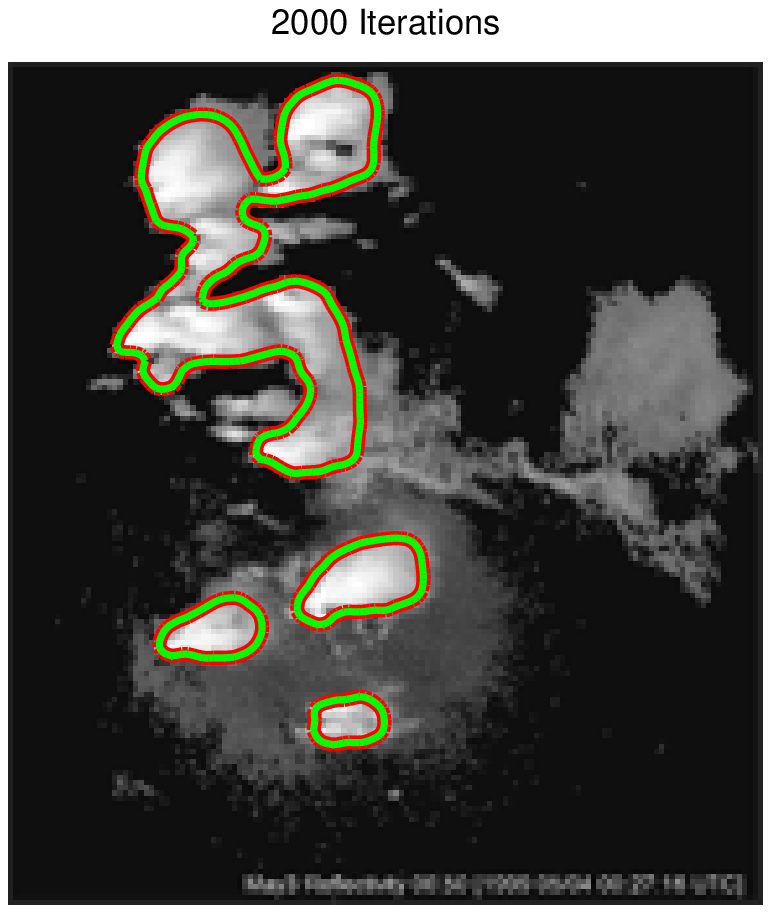}} \caption{ \bf Our model with different
$\alpha$}
\end{figure}

\end{document}